\documentclass[10pt,twocolumn,letterpaper]{article}

\usepackage{cvpr}              

\usepackage{graphicx}
\usepackage{amsmath}
\usepackage{amssymb}
\usepackage{booktabs}

\usepackage{times}
\usepackage{epsfig}
\usepackage{graphicx}
\usepackage{amsmath}
\usepackage{amssymb}
\usepackage{csquotes}
\usepackage{graphicx}
\usepackage{rotating}
\usepackage{multirow}
\usepackage[normalem]{ulem}
\usepackage{algorithm}
\usepackage{algorithmicx}
\usepackage{algpseudocode}
\usepackage{booktabs}
\usepackage{array}
\usepackage{rotating}
\usepackage{arydshln}
\usepackage{ragged2e}
\usepackage{multirow}

\usepackage{algorithm}
\usepackage{algorithmicx}
\usepackage{algpseudocode}

\usepackage{multirow}
\usepackage{tabularx}
\usepackage{makecell}
  
\usepackage{amssymb}
\usepackage{colortbl}
\usepackage{xcolor}
\usepackage{pifont}

\usepackage[normalem]{ulem}

\newcommand{\xmark}{\ding{55}}

\usepackage[pagebackref,breaklinks,colorlinks]{hyperref}

\usepackage{xcolor}
\usepackage{amssymb}
\usepackage[normalem]{ulem}
\usepackage{csquotes}

\usepackage{pifont}

\definecolor{commentcolor}{RGB}{110,154,155}   

\usepackage{newfloat}
\usepackage{listings}
\floatstyle{ruled}
\newfloat{listing}{tb}{lst}{}
\floatname{listing}{Listing}

\usepackage[capitalize]{cleveref}
\crefname{section}{Sec.}{Secs.}
\Crefname{section}{Section}{Sections}
\Crefname{table}{Table}{Tables}
\crefname{table}{Tab.}{Tabs.}

\def\cvprPaperID{8371} 
\def\confName{CVPR}
\def\confYear{2022}

\begin{document}

\title{GETAM: Gradient-Weighted Element-Wise Transformer Attention Map \\for Weakly Supervised Semantic segmentation}

\title{GETAM: Gradient-weighted Element-wise Transformer Attention Map for Weakly-supervised Semantic Segmentation}

\author{Weixuan Sun$^1$, Jing Zhang$^1$, Zheyuan Liu$^1$, Yiran Zhong$^2$, Nick Barnes$^1$\\
Australian National University$^1$    SenseTime$^2$  \\
\\
}
\maketitle

\begin{abstract}
Weakly-supervised semantic segmentation (WSSS) is
challenging, particularly when 
image-level labels are used to supervise pixel-level prediction. To bridge their gap, a Class Activation Map (CAM) is usually generated to provide pixel-level pseudo labels. 
CAMs in Convolutional Neural Networks suffer from partial activation i.e. only the discriminative regions are activated.
Transformer networks, on the other hand, are highly effective at exploring global context,
potentially alleviating the partial activation issue. 
In this paper, we introduce the Gradient-weighted Element-wise Transformer Attention Map (GETAM) and propose the first transformer-based WSSS approach.
GETAM can
show fine scale class-wise activation, revealing different parts of the object across transformer layers. 
Further, we propose an activation-aware label-completion module to generate high-quality pseudo labels. 
Finally, we incorporate the proposed methods into an end-to-end framework for WSSS using double-backward propagation. 
Extensive experiments on PASCAL VOC and COCO dataset demonstrate that our results outperform
not only the state-of-the-art end-to-end approaches by a significant margin, but also most of
the multi-stage methods.
\end{abstract}

\maketitle

\section{Introduction}
Recent work on 2D image semantic segmentation has achieved great progress via
deep fully convolutional neural networks (FCNs) \cite{long2015fully}.
The success of these models \cite{zhao2017pyramid,chen2017deeplab,chen2014semantic,chen2017rethinking} comes from large training datasets with pixel-wise labels, which are laborious and expensive to obtain. 
To relieve the labeling burden, multiple types of weak labels have been explored, including image-level labels \cite{huang2018weakly,ahn2018learning,fan2020cian}, points \cite{bearman2016s},
scribbles \cite{vernaza2017learning,lin2016scribblesup,tang2018regularized} and bounding boxes \cite{dai2015boxsup,papandreou2015weakly,lee2021bbam,oh2021background,sun20203d}.
In this paper, we focus on weakly-supervised semantic segmentation (WSSS) with image-level labels.

\begin{figure}[!t]
   \begin{center}
   {\includegraphics[width=1\linewidth]{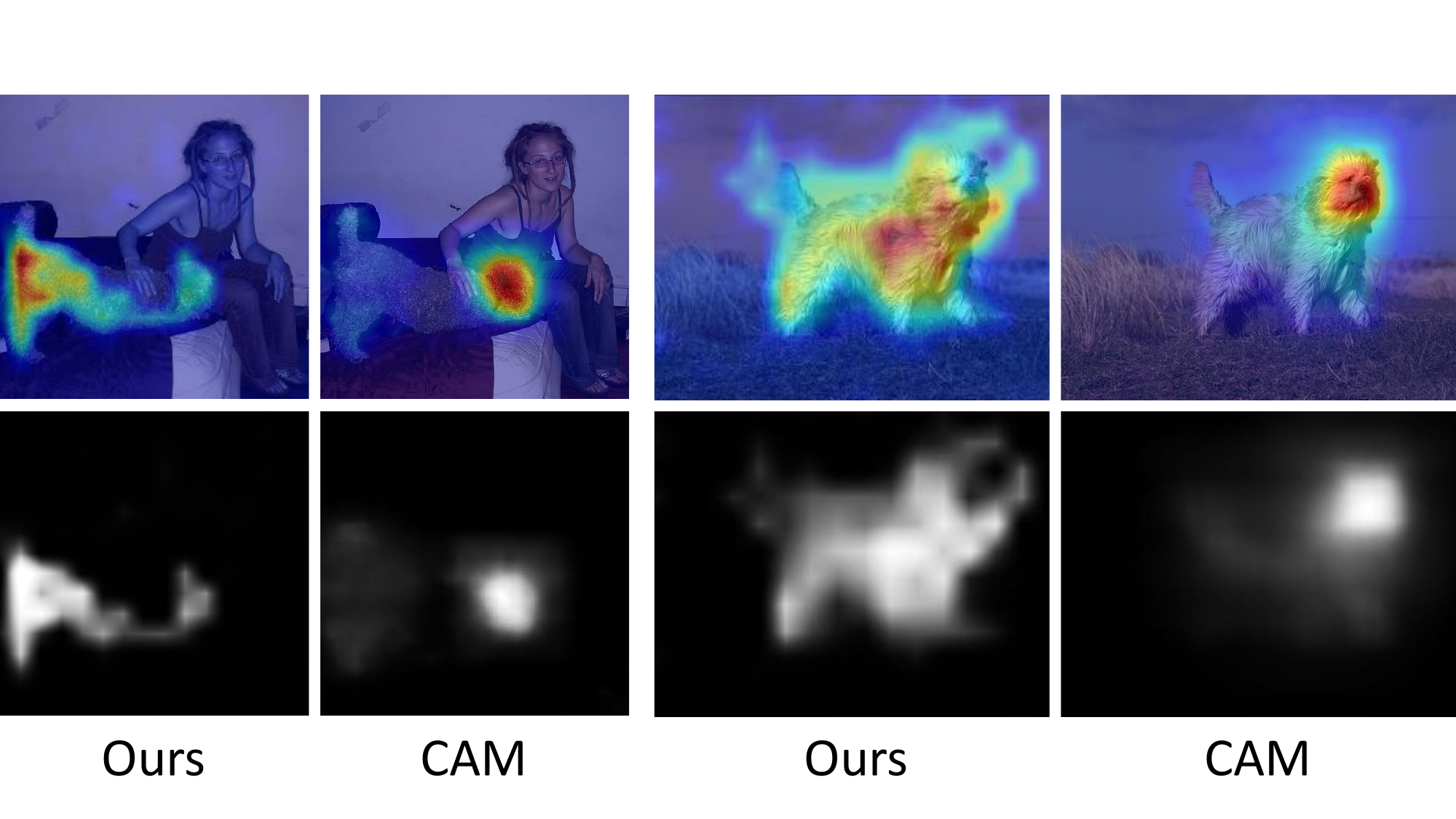}} 
   \end{center}
\caption{Comparison between CAM and our GETAM.
Baseline CAMs (CAM)
focus too much on discriminative regions and are over smoothed.
Our GETAM generates class-wise attention maps (Ours), capturing better
object shapes.}
   \label{fig:intro}
\end{figure}

To bridge the gap between pixel-level classification and image-level annotation, it is essential to localize object classes in the image from image-level labels.
Most WSSS methods rely on the Class Activation Map (CAM) \cite{zhou2016learning} as initial seeds from the image-level label to learn pixel-level labeling. 
Typical multi-stage image-level WSSS methods usually adopt progressive steps ~\cite{wang2020self,chang2020weakly,zhang2020splitting,zhang2020reliability,guo2019mixup,yun2019cutmix,kim2021discriminative}:
1) training a CNN classifier
to obtain object activation maps \cite{zhou2016learning,wang2020self,chang2020weakly,guo2019mixup,kim2021discriminative,hou2018self};
2) refining the maps with non-learning \cite{yao2021non} or learning based methods \cite{ahn2018learning,huang2018weakly}
to obtain pseudo labels; and
3) using these pseudo labels for fully-supervised training of an off-the-shelf semantic segmentation network, such as Deeplab~\cite{chen2017deeplab}.
Alternatively, recently end-to-end WSSS methods have become more prevalent \cite{zhang2020reliability,Araslanov_2020_CVPR,zhang2021adaptive}.
In both cases, the quality of the initial response map plays a key role in WSSS. However, it is recognized by many approaches \cite{wang2020self,chang2020weakly,zhang2020splitting,zhang2020reliability,guo2019mixup,yun2019cutmix,kim2021discriminative,yao2021non,sun2022inferring,wu2021embedded} that CAMs suffer from two issues: 1) CAMs tend to only activate the discriminative regions of objects; 2) the rough object activation of CAMs loses accurate object shapes.
The main causes of above issues are limited receptive fields and the progressively down-sampled CNN feature maps.

Transformers \cite{vaswani2017attention} have achieved great success in various computer vision tasks, but transformer-based WSSS is yet to be well studied.
We argue that transformers have several appealing mechanisms for WSSS. 
First, transformers are highly effective at exploring global context with long-range dependency modelling. Hence, one would expect they could address issues regarding the limited visual field of CNNs.
Also, after computing the initial embedding, vision transformers have a constant global receptive field at every layer 
with constant dimensionality, 
which could retain more accurate structure information \cite{ranftl2021vision,mao2021transformer}. Further, transformers use self-attention to spread activation across the \textit{entire} feature map rather than just discriminative regions, leading to more uniform activation.
We show that these properties are beneficial for generating object activation maps, as well for other dense prediction tasks \cite{ranftl2021vision,liu2021swin,bao2021beit,wang2021pvtv2,strudel2021segmenter}. 

Unfortunately, we observe extensive noise if we simply transplant the conventional CAM \cite{zhou2016learning} into vision transformers (Fig.~\ref{fig:pilot}). Such noise leads to poor performance and is
believed
to arise along with the global context of the transformer architecture \cite{zhang2021aggregating}.
To address this, we closely study activation and propagation of the feature maps through transformer layers. Particularly, we explore element-wise weighting to couple the attention map (i.e. $Q \times K$) with its squared gradient
to
place greater emphasis on the gradient. 
Further, as attention is progressively propagated and refined through the transformer layers, it reveals different  parts of the object.
We sum the attention maps through transformer layers, leading to more uniformly activated
object maps, as shown in Fig.~\ref{fig:intro}. We refer to this as the \textbf{Gradient-weighted Element-wise Transformer Attention Map (GETAM)}.
After obtaining class-wise attention maps,
we propose \textbf{activation aware label completion}, combining the obtained object activation with off-the-shelf saliency maps to generate pseudo segmentation labels. 
In this way, we refine foreground object masks and actively discover objects in the background, leading to high-quality pseudo labels.

Finally, we propose a \textbf{double-backward propagation} mechanism to implement our framework in an end-to-end manner. Most WSSS methods require multiple stages, involving multiple models with different pipelines and tweaks, making them hard to train and implement.
Although the existing end-to-end approaches
 \cite{pinheiro2015image,papandreou2015weakly,zeng2019joint,zhang2020reliability,Araslanov_2020_CVPR,zhang2021adaptive} 
 are elegant, they
 show substantially inferior performance to multi-stage methods.
Our method is easy to implement and extensive experiments on the PASCAL VOC~\cite{everingham2010pascal}
and COCO \cite{lin2014microsoft} datasets verify its effectiveness.

Our contributions can be summarised as:
1) We introduce the first weakly-supervised semantic segmentation framework based on vision transformer networks. The key to its performance is the Gradient-weighted Element-wise Transformer Attention Map (GETAM), capturing better object shapes than
traditional CAMs.
2) We present
activation-aware label completion guided by saliency maps to
generate high-quality pseudo labels. 
3) We propose a double-backward propagation mechanism to integrate our method in an end-to-end manner.
Despite its simplicity, our method greatly boosts the performance of single-stage WSSS, and it is the first one to be competitive with multi-stage methods.

\section{Related Work}
\subsection{Weakly Supervised Semantic Segmentation}
To save labeling cost,
various WSSS methods have been proposed,
including those using image-level labels~\cite{ahn2018learning,papandreou2015weakly,wang2020self,chang2020weakly,zhang2020reliability,yun2019cutmix,lee2021railroad}, scribbles~\cite{lin2016scribblesup}, points~\cite{bearman2016s}, and bounding boxes~\cite{dai2015boxsup,sun20203d,lee2021bbam,oh2021background}.
We mainly focus on image-level models. They
can be grouped into two families: multi-step, and
one-step
methods.
\subsubsection{Multi-step methods}
\cite{wang2020self,chang2020weakly,zhang2020splitting,zhang2020reliability,guo2019mixup,yun2019cutmix,kim2021discriminative,ahn2018learning,chang2020mixup,lee2021railroad} refine one or multiple sub-modules within the multi-model framework.
Among them, \cite{ahn2018learning} predicts semantic
affinity between a pair of adjacent image coordinates supervised by initial activation, the network is then used to guide a random walk to generate pseudo labels.
\cite{chang2020mixup,guo2019mixup} utilize Mixup data augmentation to calibrate uncertainty in prediction. They randomly mix an image pair 
to force the model to pay attention to extra regions in the image.
\cite{lee2021railroad} directly uses saliency maps to constrain object activation during classification training, so the CAMs can better follow object shapes.

\subsubsection{One-step methods}
\cite{papandreou2015weakly} adopts an expectation-maximisation mechanism, where intermediate predictions are used as segmentation labels.
\cite{zhang2020reliability} presents an
end-to-end framework to train classification and segmentation simultaneously, and integrates a method to obtain reliable segmentation pseudo labels.
\cite{Araslanov_2020_CVPR} introduces normalised Global Weighted Pooling (nGWP) to obtain better CAMs for segmentation, and adopts a Stochastic gate to encourage information sharing between deep features and shallow representations to deal with complex scenes.
With all above solutions, end-to-end WSSS is still far from being well-studied, and has clear performance gaps from multi-step methods.

\subsection{Network Visualization}
Various works have been proposed for network visualization and are leveraged for tasks like weak semantic segmentation and weak object localization.
CAM \cite{zhou2016learning} replaces the first fully-connected layer in image classifiers with a global average pooling layer to calculate class activation map.
In Grad-CAM family methods \cite{selvaraju2016grad,jiang2021layercam,chattopadhay2018grad}, the class-specific gradients flow to each feature map and methods adopt different ways to obtain a weighted sum of feature maps for visualization.
For the vision transformer, \cite{gao2021tscam} proposes to couple semantic-agnostic attention maps and semantic-aware maps for weakly supervised object localization.
\cite{chefer2021transformer,chefer2021generic} employ LRP-based relevance \cite{bach2015pixel} combined with gradients to explore the interpretability of transformer attention. However, none of above methods are specially designed for WSSS.

\begin{figure}[!t]
   \begin{center}
   {\includegraphics[width=1\linewidth]{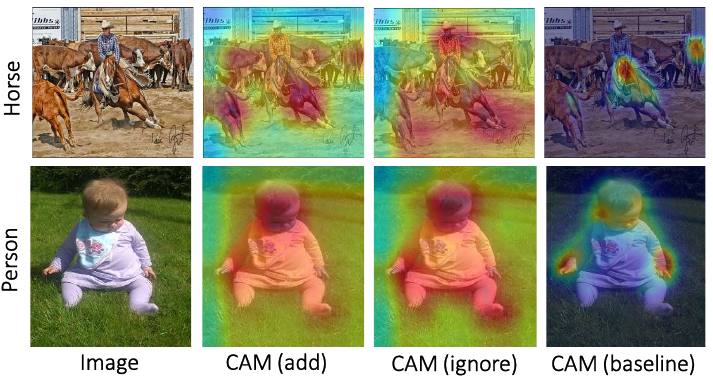}} 
   \end{center}
    \caption{Comparison between CAMs generated by ViT \textit{(add/ignore)} and CNN \textit{(baseline)},
    showing that we \textit{cannot} simply transplant the conventional CNN-based CAM into ViT.
    Specifically,
    \textit{CAM (add)}: add the \texttt{[class]} token $\boldsymbol{O}_{\text{CLS}}\in\mathbb{R}^{1\times d}$ to every location of  $\boldsymbol{O}\texttt{[1:]}\in\mathbb{R}^{n\times d}$;
    \textit{CAM (ignore)}: ignore the \texttt{[class]} token; 
    \textit{CAM (baseline)}: CNN CAM (ResNet38).
    }
    \label{fig:pilot}
\end{figure}

\begin{figure*}[!t]
   \begin{center}
   {\includegraphics[width=1\linewidth]{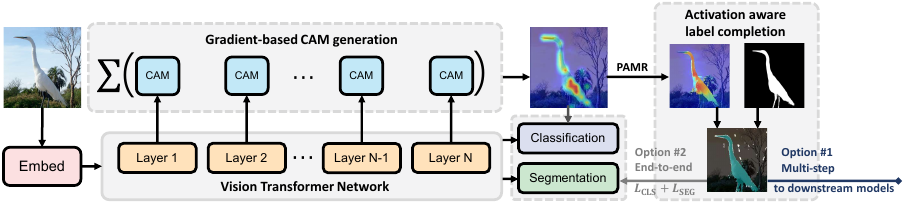}} 
   \end{center}
\caption{
Overview of the proposed end-to-end WSSS framework.
We back-propagate classification predictions, and compute the sum of class-wise attention maps from cascaded transformer blocks. 
Then we generate high-quality pixel-wise pseudo labels guided by saliency maps. 
We mainly use the pseudo segmentation labels to supervise the segmentation branch of our vision transformer in an end-to-end manner.
However, it can also be used to train a separate semantic segmentation network, as a conventional multi-stage method (see Section.\ref{sec:pseudo_label}).
}
   \label{fig:overview}
\end{figure*}

\section{Revisiting CAM in ViT}
\label{sections: pilot}
In this section, we revisit conventional CAMs \cite{ahn2018learning} used in most WSSS methods. 
We present a pilot experiment to investigate different activation mechanisms in convolutional and transformer backbones, demonstrating that the conventional CAM \cite{zhou2016learning} and Grad-CAM \cite{selvaraju2017grad} methods used in existing WSSS methods \textit{cannot} be trivially transplanted to transformer-based approaches.

To generate a CAM in a CNN network, \cite{zhou2016learning} feeds
convolutional feature maps $f$ into global average pooling (GAP) followed by a fully-connected layer to produce a categorical output.
For an image $I$, activation maps for category $c$ are defined as:
$A^c(x,y) = \theta^c_p f(I(x,y))$, 
where $\theta^c_p$ is the classifier's weight for class $c$ and  $f(I(x,y))$ is the extracted feature at location $(x,y)$.
Then, we copy the same CAM structure to the transformer. 
Given the extracted feature maps $O\in\mathbb{R}^{(n+1)\times d}$, we obtain $O\texttt{[1:]}\in\mathbb{R}^{n\times d}$, where $n = h\times w$ is the input image patch size, and $d$ is the feature embedding dimension. The first row $O_{\text{CLS}}\in\mathbb{R}^{1\times d}$ is the \texttt{[class]} token.
We explore two strategies to manipulate $O_{\text{CLS}}$ and generate a feature map $O\in\mathbb{R}^{h\times w\times d}$.
First, we add  $O_{\text{CLS}}\in\mathbb{R}^{1\times d}$ to every location of  $O\texttt{[1:]}\in\mathbb{R}^{n\times d}$, we denote this method as \textit{CAM(add)} in Fig.~\ref{fig:pilot}.
Second, we ignore  $O_{\text{CLS}}\in\mathbb{R}^{1\times d}$, and simply feed $O\texttt{[1:]}$ into linear classifier, we denote this method as \textit{CAM(ignore)} in Fig.~\ref{fig:pilot}.

As shown in Fig.~\ref{fig:pilot}, if we simply follow the conventional method, the transformer CAMs are flawed, \ie, the activation is not correctly located on targeted objects and cannot be used to generate pseudo segmentation labels.
Due to the self-attention mechanism, every feature map location encodes information from the entire image in a fully connected manner. However, classification loss is indifferent to extensive activation across objects, requiring only a sufficient pooled global average value.
So per-location features may not contribute to local classification predictions, and activation shows noise across the image, or can be completely wrong.
Reliable CAMs are crucial for WSSS, but CNN-based CAMs cannot be naively migrated into vision transformers, i.e. changing backbones of current WSSS methods to transformers is non-trivial.

\section{Approach}
We show the overview of our framework in Fig.~\ref{fig:overview}.
Firstly, we introduce GETAM (Gradient-weighted Element-wise Transformer Attention Map), which generates better class-wise attention maps with image-level labels.
Then, we introduce activation aware-label completion, which uses saliency information to produce high-quality pseudo segmentation labels from the activation maps.
Finally, we present our double-backward propagation approach to implement our method into a single-stage framework.

\subsection{GETAM}
As discussed in Sec.~\ref{sections: pilot}, conventional CAM generation methods used in CNN backbones fail to generate reliable CAMs in transformers.
Thus, inspired by Grad-CAM \cite{selvaraju2017grad}, we design a gradient-based method to obtain class-wise attention maps for vision transformers.
Note that our method is substantially different to Grad-CAM \cite{selvaraju2017grad}, albeit also based on gradients.
In Grad-CAM \cite{selvaraju2017grad}, classification predictions are back propagated to the output feature maps of the final convolutional
layer.
The obtained gradients are global average pooled, and
used to weigh neuron 
importance of the feature maps.

\begin{figure}[!t]
  \begin{center}
  {\includegraphics[width=1\linewidth]{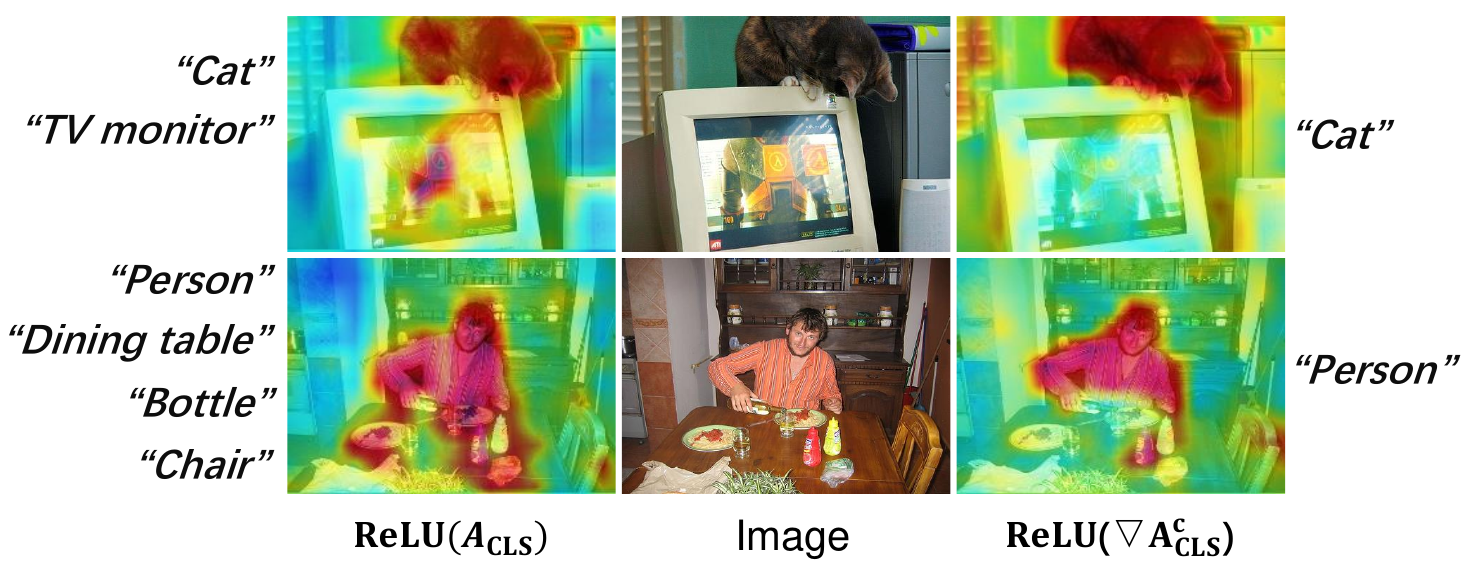}}
  \end{center}
\caption{Visualizations of class-agnostic attention maps (\textbf{leftmost column}) and class-wise gradient maps (\textbf{rightmost column}). Class-agnostic attention maps show the regions that possibly contribute to the classification predictions. Class-wise gradient maps display class specific regions and retain clear object shapes. Recommend viewing digitally. 
}
\label{fig:relu grad}
\end{figure}

\begin{figure}[!t]
  \begin{center}
  {\includegraphics[width=0.75\linewidth]{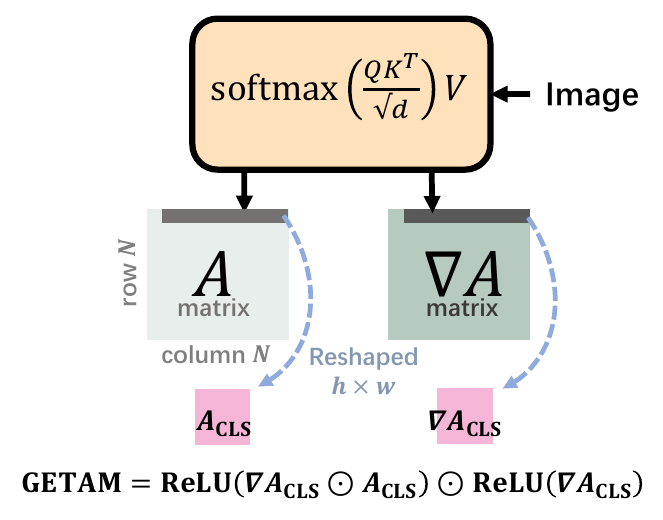}}
  \end{center}
\caption{GETAM generation in a transformer block.
$\boldsymbol{A}_{\text{CLS}}^i$: the class-agnostic attention map. $\boldsymbol{\bigtriangledown} \boldsymbol{A}_{\text{CLS}}$: the class-wise gradient map.}
\label{fig:getam}
\end{figure}

However, transformer networks use different structures for predictions. 
A transformer network consists of several successive transformer blocks, each composed of multi-head self-attention modules and feed-forward connections.
The $i^{th}$ self-attention module on one of the multiple heads in a transformer block is defined as:
\begin{equation}
\label{equation qkv}
\small
\textstyle{
\text{Attention}\left(Q^i,K^i,V^i\right) = \text{softmax}\left(\frac{Q^i {K^i}^{T}}{\sqrt{d}}\right) V^i
 = A^i V^i,}
\end{equation}
\noindent
where $Q^i,K^i,V^i$ are query, key and value matrices $\in \mathbb{R}^{(n+1)\times d}$, $n$ is number of patches ($n = w\times h$) and $d$ is feature dimensions.
The \texttt{[class]} is an extra learnable token in the first row of these matrices \cite{dosovitskiy2020image}.
As shown in Fig.~\ref{fig:getam}, $A^i\in \mathbb{R}^{n\times n}$ is the attention matrix which encodes the attention coefficient between any two positions in input images, i.e., every image token has contextual information from all other tokens.
Consequently, 
the \texttt{[class]} token attends to all token information across the image. However, it is not affected by itself as it does not represent an image location.
We define $A^i\texttt{[0,1:]} = A_{\text{CLS}}^{i} \in \mathbb{R}^{n} $.
We can reshape $A_{\text{CLS}}^{i} \in \mathbb{R}^{n}$ back to the image shape $A_{\text{CLS}} \in \mathbb{R}^{h\times w}$, and obtain a class-agnostic attention map, where every position in the map denotes its contribution to classification.
In every transformer block with multiple self-attention heads, we define $A_{\text{CLS}}^i$ as the average across all heads for simplicity. 
As empirically verified in \cite{gao2021tscam}, $A^I_{\text{CLS}}$
aggregates attention from all heads to display the object extent that possibly contributes to classification predictions (see the leftmost column
of Fig~\ref{fig:relu grad}).

As $A_{\text{CLS}}^{i}$ is class agnostic, we couple it with its gradients
$\bigtriangledown A^{i}$ to obtain the class-wise attention map.
We back-propagate the classification
score $y^c$ for each class $c$.
The graph is differentiated using the chain rule through the transformer network.
In the $i^{th}$ block, 
the gradient map $\bigtriangledown A^{i,c}$ is obtained with respect to the attention matrix $A^i$.
Referring to Fig.~\ref{fig:getam}, we extract the corresponding gradient map
$\bigtriangledown A_{\text{CLS}}^{i,c}\in \mathbb{R}^{n} $ of $A_{\text{CLS}}^i$ with respect to class $c$.
As discussed in~\cite{selvaraju2017grad,chattopadhay2018grad,jiang2021layercam}, a positive gradient corresponding
to a location in the feature map indicates that it has a positive influence on
the prediction score of the target class.
We find that this assertion still holds in vision transformers.
Each position in $\bigtriangledown A_{\text{CLS}}^{i,c}$ indicates the contribution of this token to the classification output of class $c$.

\subsubsection{Attention Gradient Coupling}
We observe that
$A_{\text{CLS}}^i$ shows class-agnostic areas with possibly targeted objects and relatively clean background. 
Further, we find $\bigtriangledown A_{\text{CLS}}^{i,c}$ is noisy but retains more complete and clear object shapes in comparison (see rightmost column of Fig.~\ref{fig:relu grad}). 

Based on the above observations, we propose to combine the attention map and its gradient inside every transformer block to generate reliable class-wise attention as shown in Fig.~\ref{fig:getam}.
Formally, the class-wise attention map of block $i$ for class $c$ is defined as:
\begin{equation}
\small
\text{GETAM}_c^i = \text{ReLU}\left(\bigtriangledown A_{\text{CLS}}^{i,c} \odot A_{\text{CLS}}^{i}\right) \odot \text{ReLU}(\bigtriangledown A_{\text{CLS}}^{i,c})
\end{equation}
\noindent
We first use an element-wise multiplication $\odot$ to couple the attention map $A_{\text{CLS}}^{i}$ with its gradient map $\bigtriangledown A_{\text{CLS}}^{i,c}$.
Then, we perform another element-wise multiplication with $\text{ReLU}(\bigtriangledown A_{\text{CLS}}^{i,c})$. 
Different from Grad\_CAM which uses a Global Average Pool of the Gradients, 
our attention map is weighted element-wise by the square of the gradient, and negative responses of attention and gradients are eliminated by the ReLUs. 
Our Attention-Gradient coupling can better harvest spatial locations that have positive contributions to the targeted class, while suppressing noisy regions. Squaring places greater emphasis on the gradient as it shows more complete object shapes than the CAMS.

\subsubsection{Successive Attention Aggregation}
\begin{figure}[!t]
   \begin{center}
   {\includegraphics[width=1\linewidth]{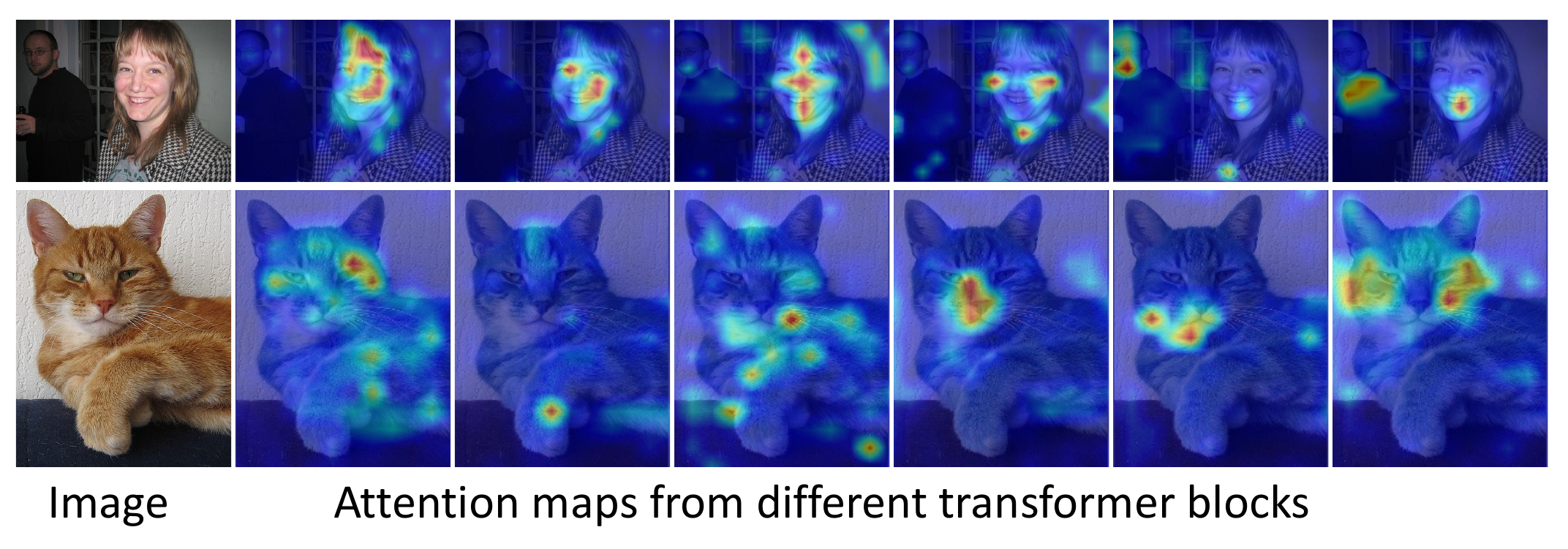}} 
   \end{center}
\caption{Sample attention maps from successive transformer blocks of our trained model in one forward pass. The cascaded maps focus on different object regions without low-level noise.
In the first row, the network focuses on different people in the foreground and background separately at different layers.
Recommend viewing digitally. 
}
\label{fig:combine}
\end{figure}

After obtaining the class-wise attention map in a single transformer block, here we present an analysis on how to aggregate class attention maps from cascaded transformer blocks. 
As commonly recognized by existing WSSS methods~\cite{sun2022inferring,Sun_2021_ICCV,Sun_2021_ICCV,Zhang_2021_ICCV,wang2020self}, the attention maps should not respond too sparsely (i.e., only highlighting discriminative regions), nor be overly smoothed.
Based on the above requirements, we first visualize the class-wise attention maps from different transformer blocks in Fig.~\ref{fig:combine}.
Unlike CNNs where low-level features contain too much noise that buries useful information \cite{wei2021shallow,jiang2021layercam}, we observe that the cascaded maps in vision transformers tend to focus on discriminative regions. For instance, in Fig.~\ref{fig:combine} (bottom), attention maps from different layers reveal different regions of the cat (e.g., cheek, nose, chin, body and hand).
Thus, it is crucial to choose an appropriate fusion method to combine different class-wise attentions maps.

\begin{figure}[!t]
   \begin{center}
   {\includegraphics[width=0.75\linewidth]{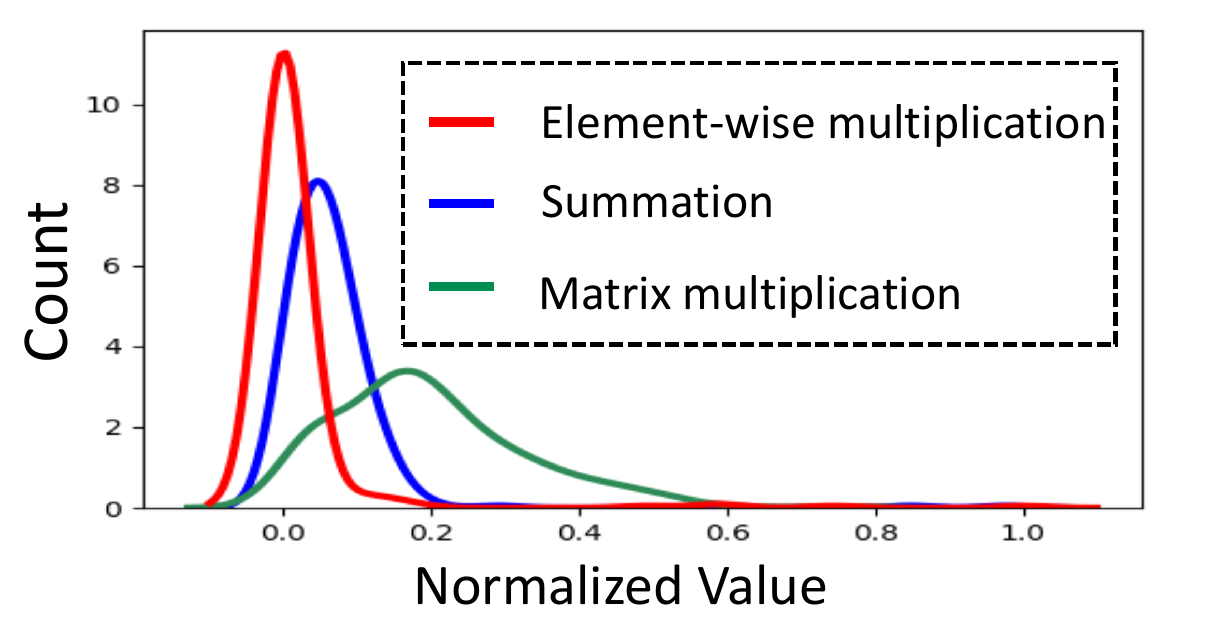}} 
   \end{center}
\caption{Distributions of class-wise attention maps using different fusion methods. Element-wise multiplication concentrates the distribution, but suppresses most areas. Matrix multiplication (dot production) smooths the attention, leading to additional noises.
}
\label{fig:distribution}
\end{figure}

In this view, we present a numerical analysis of different fusion approaches (Fig.~\ref{fig:distribution}). 
On the PASCAL VOC \cite{everingham2010pascal} training set, we apply three commonly adopted operations (element-wise multiplication, summation and matrix multiplication) to fuse the successive maps, and then visualize the distributions of the fused results after normalization.
As illustrated in Fig.~\ref{fig:distribution}, element-wise multiplication will concentrate the values and most areas are suppressed, since the activation is canceled if the value is low 
in any level of the transformer. Contrarily, matrix multiplication (dot production) will smooth the attention, leading to additional noise in non-object regions.
Based on the above analysis, we propose summation aggregation across cascaded attention maps, which encourages the final class-wise attention maps to cover accurate object areas and does not over-smooth them.
Formally, the attention maps are added through $L$ layers of the vision transformer:
\begin{equation}
    \text{GETAM}_c = \sum^L_{i} (\text{GETAM}_c^i)
    \label{SumGETAM}
\end{equation}
Fig.~\ref{fig:combine} shows that GETAM from the cascaded blocks (Eq.~\ref{SumGETAM})
captures reliable object shapes and suppresses noise.
See supplementary material for 
qualitative comparison.

\subsection{Activation Aware Label Completion}
GETAM generates reliable class-wise attention maps.
However, the activation maps require refinement, so they can be served as pseudo segmentation labels.
To achieve this, 
we first adopt pixel adaptive mask refinement (PAMR) \cite{Araslanov_2020_CVPR},
a parameter-free recurrent module that efficiently refines pixel labels using local information. 
Then, we propose
\textit{saliency constrained object masking} and \textit{high activation object mining} to obtain high-quality pseudo 
labels. 
The two solutions work collaboratively to
support accurate segmentation of salient objects, but do not suppress non-salient objects. 

\subsubsection{Saliency Constrained Object Masking}
We observe that due to the global context of self-attention, candidate regions that may contribute to classification
are activated, leading to high recall of our activation maps for targeted objects.
Saliency maps from off-the-shelf models can provide precise foreground object shapes and have been used as background cues to many WSSS approaches \cite{fan2020learning,jiang2019integral,lee2019ficklenet,sun2020mining,wang2018weakly,lee2021railroad,yao2021non}. 
Using a novel approach, we leverage 
the saliency mask to constrain activated object regions, which is particularly necessary in our case for GETAM to combat additional activation on object boundaries.

\begin{figure}[!t]
   \begin{center}
   {\includegraphics[width=1\linewidth]{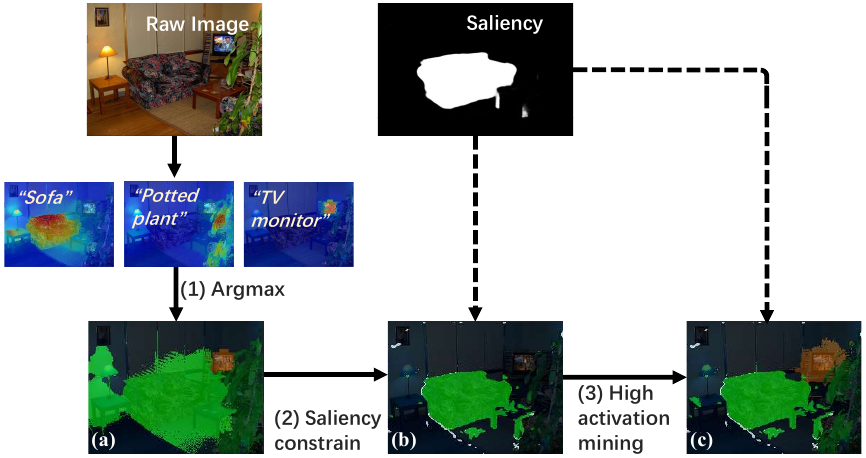}} 
   \end{center}
\caption{Activation aware label completion module.
\textbf{(a)} First locate all objects
with activation maps.
\textbf{(b)} With the saliency constraint, we obtain accurate shape for salient objects, however,
non-salient objects are suppressed.
\textbf{(c)} High activation mining correctly locates suppressed background objects.
Enlarge the figure by 3 times for best viewing.
}
   \label{fig:pseudo generation}
\end{figure}

First, based on object activation maps $M_{\text{fg}} = \text{GETAM} \in \mathbb{R}^{C\times h\times w }$ which have $C$ foreground classes, 
we calculate an arbitrary background channel in a similar way to \cite{ahn2018learning}: 
$
M_{\text{bg}} = \left[ 1-\max_{c\in{C_{\text{fg}}}}\left(M_{\text{fg}}^{c}(i)\right) \right] ^{\gamma}
$, 
where $\gamma>1$ is a parameter to adjust background labels and
$i$ is the pixel position.
Then we concatenate 
$M_{\text{bg}}$ onto $M_{\text{fg}}$ to form activation maps $M$,
and use $\text{argmax}(M)$ to find the per-pixel highest activation.
After that, we can locate all possible objects in both salient and non-salient regions with rough boundaries, as shown in Fig.~\ref{fig:pseudo generation} (a).
Then we adopt saliency maps 
to refine object boundaries of $M$, where
the current temporary background (0) is set to unknown (255) and
all non-salient regions 
are set to background (0).
This is based on the observation
that the saliency maps normally provide accurate object boundaries \cite{yao2021non,lee2021railroad,wang2021weakly}. 
Formally, consider activation maps $M$ with saliency map $S$, then pixel $i$ of our pseudo label $P_\text{seg}$ is:
\begin{equation}
P_{\text{seg}}(i) = \begin{cases}
     c  &  \text{$\text{argmax}(M) = c$,  $S(i) = 1$} \\
    255   &  \text{$\text{argmax}(M) = 0$,  $S(i) = 1$} \\
    0   &  \text{$S(i) = 0$}
 \end{cases}. 
\end{equation}

\subsubsection{High Activation Object Mining} With saliency constrained object masking, we obtain $P_{\text{seg}}$, where the structure of the instances that are consistent with salient objects is refined. 
However, recall that existing salient object detection models are trained with class-agnostic objects and center bias, and so
non-salient objects may masked as background. 
As shown in Fig.~\ref{fig:pseudo generation}(b), the TV monitor and potted plant are mislabeled as background. 
We then propose a high activation object mining strategy to solve this issue.

GETAM can correctly locate all desired objects in both salient and non-salient regions as shown in Fig.~\ref{fig:pseudo generation}(a).
For class $c$, 
we find high confidence regions by searching for pixels with activation greater than a threshold $\alpha$ in non-salient regions. 
We treat these
as pseudo labels of class $c$ in the background (see Fig.~\ref{fig:pseudo generation}(c)). 
In addition, we maintain another high-confidence conflict mask $M_{\text{conflict}}$ to register conflict areas in non-salient regions. That is,
if a pixel is highly activated by more than one class in the background, we regard it as conflict and label it as unknown (255) to avoid introducing incorrect labels. 
Formally, high activation object mining is defined:
\begin{equation}
P_{\text{seg}}(i) = \begin{cases}
     c  &  \text{$M(i) > \alpha$, $M_{\text{conflict}}(i) = 1$,   $S(i) = 0$} \\
    255   &  \text{$M(i) > \alpha$, $M_{\text{conflict}}(i) > 1$,   $S(i) = 0$} \\
    0   &  \text{$M(i) < \alpha$, $S(i) = 0$}
 \end{cases},
\end{equation}
where $\alpha$ is the high confidence threshold, empirically set to 0.9, i.e., if the activation at a pixel of class $c$ is higher than $90\%$ of all activation of the same object class, we regard it as highly activated and label it as $c$, otherwise as background (0). 
We ablate $\alpha$ in detail in the supplementary material.

We give a quantitative comparison of our pseudo labels to other methods in Table \ref{table pseudo label}. Our high-quality pseudo labels can be directly used to supervise an off-the-shelf semantic segmentation model. 
They can also supervise a segmentation branch of the same vision transformer backbone in an end-to-end manner (detailed below).

\begin{figure}[!t]
   \begin{center}
   {\includegraphics[width=1\linewidth]{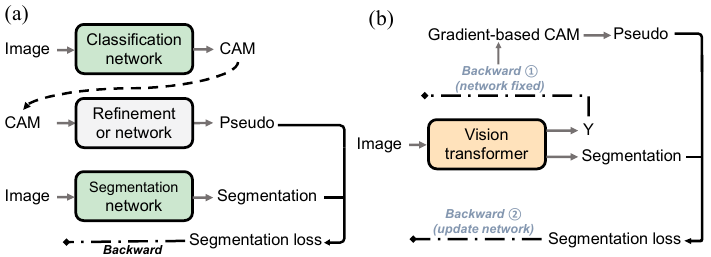}} 
   \end{center}
\caption{WSSS framework conceptual comparison: \textbf{(a)} Multi-step methods, which involve multiple stages and networks; \textbf{(b)} Proposed transformer approach, can be trained in one go. 
}
\label{fig:double bakcward}
\end{figure}

\subsection{Single-stage Double-backward Propagation}

Most current WSSS methods~\cite{wang2020self,chang2020weakly,zhang2020splitting,zhang2020reliability,guo2019mixup,yun2019cutmix,kim2021discriminative} require multiple steps, generally involving
training 
multiple models with
different pipelines and tweaks.
Inter-dependencies between the steps can easily influence final performance.
However,
the vision transformer shows properties that are especially advantageous for dense prediction tasks like semantic segmentation \cite{ranftl2021vision,liu2021swin,bao2021beit,wang2021pvtv2}. 
Hence, 
we propose an unified framework to train our WSSS method in an end-to-end manner.
As shown in Fig.~\ref{fig:overview},
the framework has two parallel branches, i.e. a classification and a semantic segmentation branch. 
Both branches share the same vision transformer backbone, and update the entire network simultaneously during training.



As illustrated in Fig.~\ref{fig:double bakcward}, 
the core of our approach is double-backward propagation. 
That is, the network back propagates to compute GETAM without updating to obtain pseudo labels, then back propagates again to optimize the network supervised by these pseudo labels. 
Each iteration consists of two back propagation operations but the network is only updated once.
Specifically, we first perform a forward pass to produce classification predictions $Y$ without calculating the classification loss.
Then, for the target class $c$, we back propagate its output $y^c$ to obtain GETAM.
We iterate to obtain class-wise attention maps for every class appearing in the image, and use them
to generate pseudo segmentation labels.
In the second back propagation step, we clear the gradients and perform another forward pass to generate classification and semantic segmentation predictions. With the pixel level pseudo labels and image level labels, we train our network with classification and segmentation predictions from the second back propagation.
Back propagating GETAM
leads to improved localization of objects and segmentation performance. 
The proposed double-backward propagation does not rely on
multi-step WSSS training, and shows competitive performance (see Table \ref{table sota}).

\section{Experiments}
\subsection{Setup and Implementation Details}
Our method is evaluated on the PASCAL VOC \cite{everingham2010pascal} and MS-COCO datasets \cite{lin2014microsoft}.
PASCAL VOC \cite{everingham2010pascal} has one background and 20 foreground classes. The official dataset consists of 1,446 
training, 1,449 validation and 1456 test images.
We follow common practice \cite{hariharan2011semantic}, augmenting the training set to form a total of 10,582 images.
MS-COCO 2014 \cite{lin2014microsoft} contains 81 classes including the background class with 80k train and 40k val images and more complex scenes, which is more difficult for WSSS.
In addition, we adopt \cite{mao2021transformer} as our saliency detection model by re-implementing it to generate saliency maps on both datasets.
The backbone network of our end-to-end framework is ViT \cite{dosovitskiy2020image} 
. 
We also test our method on other backbones including ViT-Hybrid \cite{dosovitskiy2020image}, Deit \cite{touvron2021training} and Deit\_Distilled \cite{touvron2021training}.


Our end-to-end framework has two branches, \ie~a classification branch and segmentation branch, and they share the same vision transformer backbone.
For the segmentation branch, we adopt the decoder from
\cite{ranftl2021vision}.
We train our network for 20 epochs, separated into two stages.
In the first 10 epochs we only update the classification branch in order to generate reliable class attention maps.
In the remaining 10 epochs we switch on the segmentation branch and simultaneously optimize two branches.  
Further, our generated pseudo labels can also be used in a multi-stage fashion, we use the pseudo labels to train Deeplab v2 \cite{chen2017deeplab} with the ResNet-101 backbone.
Our model is implemented with PyTorch, with reproduction details in the supplementary material.

\subsection{Ablation Studies}
\subsubsection{Effectiveness of GETAM}
The proposed GETAM is a class-wise activation visualization method for vision transformers.
We explore different ways of generating CAMs on transformer backbones and report their performances on PASCAL VOC
as shown in Table~\ref{table ablation cam}.
First, we directly leverage CAM \cite {zhou2016learning} and Grad\_CAM \cite{selvaraju2016grad} on vision transformers as baselines.
The proposed GETAM achieves significant improvements over CAM and Grad\_CAM (+16.4 mIoU and +9.2 mIoU), demonstrating that it is non-trivial to directly use existing visualization methods on the transformer networks.
TS\_CAM \cite{gao2021tscam} is a transformer based localization method, we follow its implementation and report the segmentation result.
Furthermore, in GETAM we propose to couple the attention maps with the square of corresponding gradients, 
we implement a variation that we directly compute the element-wise production of the attention maps and gradient maps, we denote this method as  
$Att \odot Gradients$.
Our GETAM achieves a 2.6 mIoU improvement over this variation. It verifies the effectiveness of the coupling strategy of GETAM.

\begin{figure}[htb]
   \begin{center}
   {\includegraphics[width=0.95\linewidth, height=55mm]{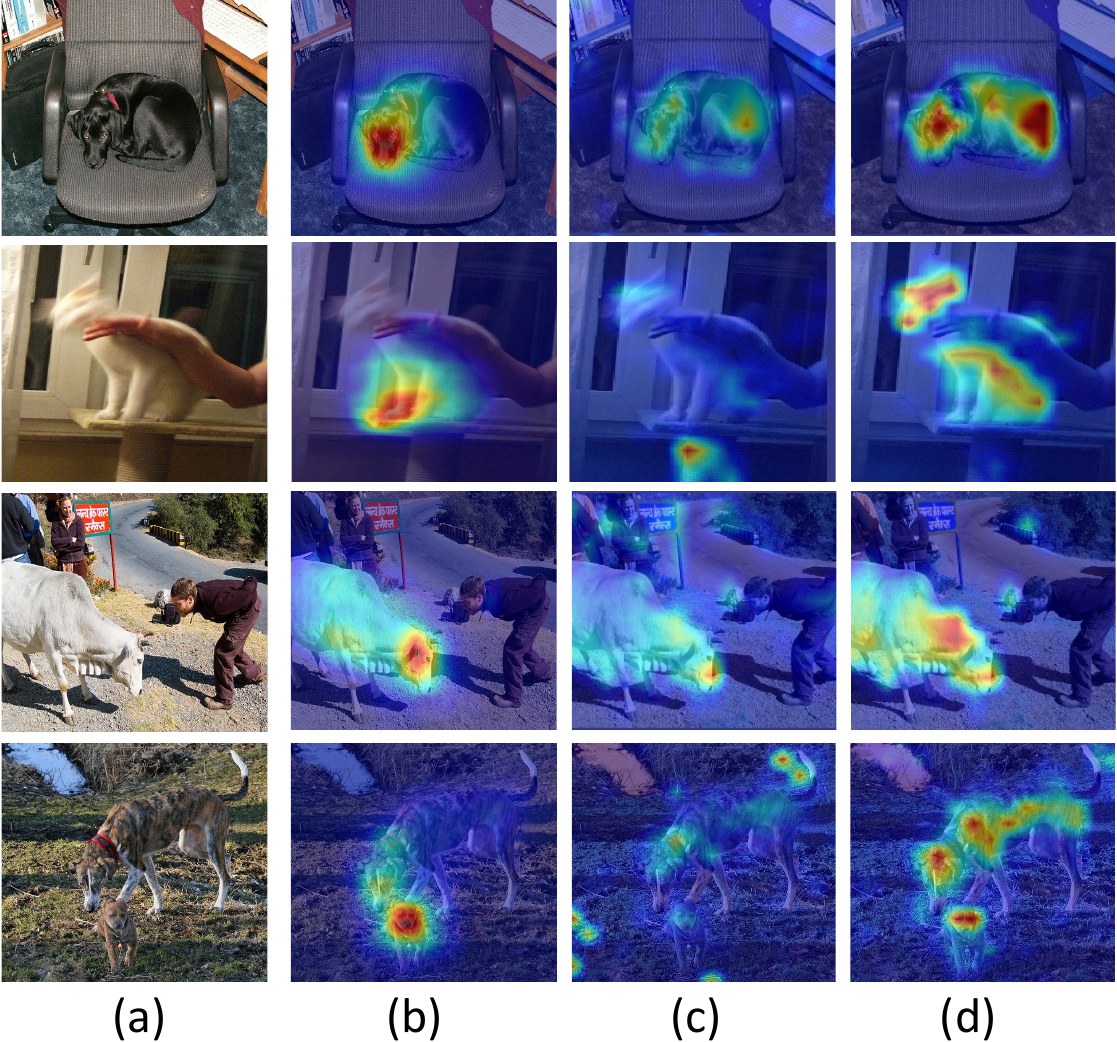}} 
   \end{center}
\caption{Visualization examples of activation maps from different methods. (a) Image; (b) CNN-based CAM~\cite{zhou2016learning}; (c) ViT-based Grad-CAM~\cite{selvaraju2016grad};  (d) GETAM. GETAM can
better capture object shapes and suppress noise than regular CAMs.}
\label{fig:cam}
\end{figure}

\begin{table}
\footnotesize
\centering 
\setlength{\tabcolsep}{4.3mm}
\begin{tabular}{cccr}
    \toprule
    Method & Backbone  & {\makecell[c]{CAM\\ generation}} &  mIoU(\textit{val}) \\
    \midrule
    & CNN  & CAM       &    64.7             \\ \hline
    & ViT & TS\_CAM\cite{gao2021tscam} & 49.7 \\
    & ViT    & CAM        &             55.3                \\
    & ViT    & Grad\_CAM          &         62.5             \\
    & ViT    & $Att \odot Gradients$          &         69.1           \\
    \textbf{Ours} & ViT    & \textbf{GETAM}   &              71.7        \\
  \bottomrule
\end{tabular}
\caption{Ablation of different CAM generation methods. 
$\boldsymbol{Att \odot Gradients}$ denotes directly computing the element-wise production of the attention maps and gradients.
}
  \label{table ablation cam}
\end{table}



\subsubsection{Effectiveness of Activation Aware Label Completion}
We propose activation aware label completion to generate pseudo labels from CAMs.
In Table \ref{table ablation pseudo}, 
we demonstrate that the proposed Activation Aware Label Completion brings improvements over the CRF-based method under two settings. 
For the CNN network it improves the baseline \cite{zhang2020reliability} by 2.1 mIoU,
for the ViT network the improvement is 7.7 mIoU.
This validates that our activation aware label completion helps generate better pseudo labels from the activation maps on both CNN and transformer backbones.

\begin{table}
\footnotesize
\centering 
\setlength{\tabcolsep}{3.3mm}
\begin{tabular}{ccccr}
    \toprule
    Model  & Backbone   & {\makecell[c]{CAM\\ generation}} & \makecell[c]{Pseudo\\generation}  & mIoU(\textit{val}) \\
    \midrule
    RRM\cite{zhang2020reliability} & CNN  &  CAM   & CRF    &    62.6 \\
     &    CNN  & CAM & \textbf{AA}       &    64.7             \\ 
    \hline
          & ViT    & \textbf{GETAM}  &   CRF    &           64.0               \\
    \textbf{Ours}     & ViT     & \textbf{GETAM}   &     \textbf{AA}     &            71.7          \\
    \bottomrule
    
\end{tabular}
\caption{Ablation of pseudo label generation methods. We use bold to indicate our proposed components.
\textbf{AA}: activation aware label completion module.
It shows that activation aware label completion improves results on both CNN and ViT backbones.}
\vspace{-5mm}
\label{table ablation pseudo}
\end{table}

\begin{table}[t!]
\footnotesize
\centering\scalebox{1}{
\setlength{\tabcolsep}{7.5mm}
\begin{tabular}{lrrrr}
\toprule
Methods & Train & Val \\
\midrule
\textit{Multi stage methods}  \\
\hline
SEAM\cite{wang2018weakly} + RW + CRF& 63.6 & --  \\
SC\_CAM \cite{chang2020weakly} + RW + CRF&  63.4 & 61.2\\
SEAM + CDA\cite{Sun_2021_ICCV} & 66.4 & --\\
CAM + RW + IRN\cite{ahn2019weakly}  &  66.5     &      --      \\
IRN + CDA\cite{Sun_2021_ICCV} & 67.7 & -- \\
CPN \cite{Zhang_2021_ICCV} & 68.0 & -- \\
EDAM \cite{wu2021embedded} + CRF & 68.1 & --\\
AdvCAM \cite{lee2021anti} + RW + CRF & 68.0 & --\\
AdvCAM \cite{lee2021anti} + IRN + CRF & 69.9 & --\\
\midrule
\textit{Single stage methods}  \\
\hline
1-stage-wseg\cite{Araslanov_2020_CVPR}  &   64.7         & 63.4             \\
1-stage-wseg\cite{Araslanov_2020_CVPR} + CRF  &   66.9         & 65.3             \\
Ours   & 69.6         & 66.7             \\
\bottomrule
\end{tabular}
}
\caption{Pseudo label mIoU on PASCAL VOC \textit{train} and \textit{val} set. Our results are obtained on ViT-Hybrid~\cite{dosovitskiy2020image}.
}
\label{table pseudo label}
\end{table}

\subsubsection{Pseudo Label Quality}
\label{sec:pseudo_label}
Pseudo label quality is crucial to segmentation performance for WSSS. 
We extract the generated pseudo labels in our end-to-end training process and evaluate them for quality with the PASCAL VOC ground-truth.
The results in Table \ref{table pseudo label} show that 
the pseudo segmentation labels of GETAM outperform all end-to-end methods. 
Further, recent state-of-the-art multi-step approaches focus on obtaining pseudo labels using sophisticated pipelines and training multiple networks, our pseudo labels are still comparable to the best of these. 
In addition, we trained deeplabv2 \cite{chen2017deeplab} using our pseudo labels in an multi-stage manner
and achieve a 70.6 mIoU on the PASCAL validation set.
We report single-stage methods in Table \ref{table sota}.

\begin{table}[]
\footnotesize
\centering \scalebox{0.9}{ 
\setlength{\tabcolsep}{7mm}
\begin{tabular}{lcc}
\toprule
Backbone & MIoU (Val) & MIoU (Test) \\
\midrule
AALR\cite{zhang2021adaptive} (ACMMM2021) & 63.9 & 64.8 \\
\midrule
ViT &  68.1 & 68.8 \\
ViT-Hybrid & 71.7 & 72.3  \\
Deit &  66.0 &  68.9 \\
Deit-Distilled &  70.7 & 71.1 \\
\bottomrule
\end{tabular}
}
\caption{
Performances on different vision transformers. Our method performs consistently better.
}
\label{table backbones}
\end{table}

\begin{figure*}[tb!]
   \begin{center}
   {\includegraphics[width=0.95\linewidth]{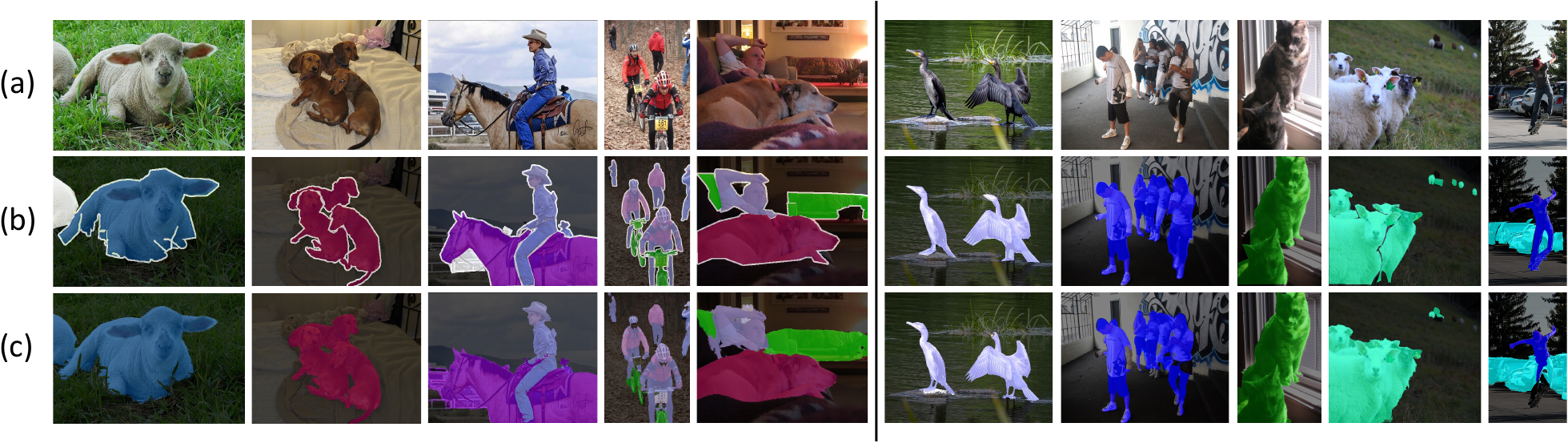}} 
   \end{center}
\caption{Qualitative segmentation results on the PASCAL VOC and MS COCO. (a) Image; (b) Ground-truth; (c) Ours.
}
\label{fig:seg pred}
\end{figure*}

\subsubsection{Results with Different Vision Transformer Backbones}
We investigate different vision transformer backbones including ViT, ViT-Hybrid \cite{ranftl2021vision}, Deit and Deit-Distilled \cite{touvron2021training} in our end-to-end WSSS framework, where all networks have 12 transformer layers 
. In ViT\_Hybrid, we aggregate attention maps from last 6 layers so as to decrease low-level noise from convolution, while in other backbones we aggregate attention from all 12 layers \footnote{We refer the reader to the supplementary material for ablation.}. 
As reported in Table \ref{table backbones}, our method performs consistently better than the previous state-of-the-art single-stage method AALR \cite{zhang2021adaptive}, validating the effectiveness of our approach.
The efficiency of the proposed GETAM on various backbones also makes it a visualization tool for transformer networks(Fig.~\ref{fig:cam}).
It is widely recognized that convolutional structures in transformers provide performance gains in vision tasks \cite{dosovitskiy2020image,wu2021cvt,dai2021coatnet}. Here we can see that ViT\_Hybrid achieves the best performances among these backbones.

\begin{table}[t]
\footnotesize
\centering \scalebox{0.75}{
\setlength{\tabcolsep}{2.5mm}
\begin{tabular}{llrrrr}
\toprule
&Method & Backbone & Sup. & val & test \\
\midrule
\multirow{15}{*}{\rotatebox{90}{Multi-stage }}
& SEAM\cite{wang2020self} (CVPR2020)  & ResNet38 & I      &  64.5   &     65.7 \\   
&SC-CAM \cite{chang2020weakly} (CVPR2020)    &    ResNet101    &    I   &  66.1   &  65.9      \\   
&CONTA\cite{zhang2020causal} (NeurIPS2020) &ResNet38    &  I        &  66.1   &  66.7\\
&CDA \cite{Sun_2021_ICCV} (ICCV2021) &ResNet101    &  I        &  66.1   &  66.8\\
&MCS \cite{sun2020mining} (ECCV2020)    & ResNet101&  I+S    &  66.2   &  66.9      \\   
&ECS-Net\cite{Sun_2021_ICCV} (ICCV2021) & ResNet38 &  I+S    &  66.6   &  67.6      \\
&EME\cite{fan2020employing} (ECCV2020)      & ResNet101     &   I+S &67.2    &  66.7   \\
&ICD\cite{fan2020learning} (CVPR2020)  &     ResNet101     &   I+S       & 67.8    &  68.0\\
&CPN\cite{Zhang_2021_ICCV} (ICCV2021) & ResNet101     &     I    & 67.8   &  68.5\\
&CGNet\cite{Kweon_2021_ICCV} (ICCV2021) & ResNet38 & I & 68.4 & 68.2 \\
&AuxSegNet \cite{Xu_2021_ICCV} (ICCV2021) & ResNet101     &     I+S    & 69.0    &  68.6\\
&PMM \cite{Li_2021_ICCV} (ICCV2021) & ResNet101    &     I   & 70.0    &  70.5\\
&RIB\cite{lee2021reducing}(NeurIPS2021) & ResNet101 & I+S & 70.2 & 70.0 \\
&NSRM\cite{yao2021non} (CVPR2021)      & ResNet101     &      I+S   & 70.4    &  70.2\\
&DRS\cite{kim2021discriminative} (AAAI2021) & ResNet101 & I & 70.4 & 70.7 \\
&VWL-L\cite{ru2022weakly} (IJCV2022) & ResNet101 & I & 70.6 & 70.7 \\
& EDAM\cite{wu2021embedded} (CVPR2021) & ResNet101 & I+S & 70.9 & 70.6 \\
& EPS\cite{lee2021railroad}(CVPR2021) & ResNet101 & I+S & 71.0 & 71.8 \\
&URN\cite{li2021uncertainty} (AAAI2022) &ResNet101 & I & 71.2 & 71.5 \\
\midrule
\multirow{8}{*}{\rotatebox{90}{Single-stage }}
&EM \cite{papandreou2015weakly} (ICCV2015) & VGG16 & I &    38.2 & 39.6 \\
&TransferNet\cite{hong2016learning} (CVPR2016) & VGG16 & I+COCO & 52.1 & 51.2 \\
&CRF-RNN \cite{roy2017combining} (CVPR2017) & VGG16 & I  &  52.8 &53.7 \\
&RRM\cite{zhang2020reliability} (AAAI2020)     &  ResNet38   & I         & 62.6   &  62.9   \\
&1-stage-wseg\cite{Araslanov_2020_CVPR} (CVPR2020)  &  ResNet38   & I          & 62.7   &  64.3   \\
&JointSaliency\cite{zeng2019joint} (ICCV2019) & DenseNet169 & I+S & 63.3 & 64.3 \\
&AALR\cite{zhang2021adaptive} (ACMMM2021) & ResNet38 & I & 63.9 & 64.8 \\
\rowcolor{lightgray} &GETAM(ours) & ViT-Hybrid & I+S &  \textbf{71.7}  & \textbf{72.3}  \\
\bottomrule
\end{tabular}
}
\caption{Comparison with the state-of-the-art methods on PASCAL VOC 2012 \textit{val} and \textit{test} sets. 
Different supervision is used: I: image-level label. COCO: MS-COCO~\cite{lin2014microsoft}, S: saliency.
}
\label{table sota}
\end{table}

\begin{table}[t]
\footnotesize
\centering \scalebox{0.8}{
\setlength{\tabcolsep}{2.5mm}
\begin{tabular}{lrrr}
\toprule
Method  & Backbone &  Single-stage & MIoU (val) \\
\midrule
SEC \cite{kolesnikov2016seed} (CVPR2016) & VGG16  &\xmark & 22.4\\
DSRG  \cite{huang2018weakly} (CVPR2018) &  VGG16 &\xmark& 26.0\\
Wang \cite{wang2018weakly} (IJCV2020) & VGG16 &\xmark& 27.7\\
Luo \cite{luo2020learning} (AAAI2020) & VGG16 &\xmark& 29.9\\
SEAM\cite{wang2020self} (CVPR2020) & ResNet38 &\xmark& 31.9 \\
CONTA\cite{zhang2020causal} (NeurIPS2020) & ResNet38 &\xmark& 32.8 \\
CDA \cite{Sun_2021_ICCV} (ICCV2021) & ResNet50 &\xmark& 33.7\\
AuxSegNet \cite{Xu_2021_ICCV} (ICCV2021) & ResNet101 &\xmark& 33.9 \\
EPS \cite{lee2021railroad} (CVPR2021) & ResNet101 &\xmark& 35.7 \\
VWL-L \cite{ru2022weakly} (IJCV2022) & ResNet101 & \xmark & 36.2 \\
CGNet\cite{Kweon_2021_ICCV} (ICCV2021) & ResNet38 & \xmark & 36.4 \\
PMM \cite{Li_2021_ICCV} (ICCV2021) & ResNet38 &\xmark& 36.7 \\
URN\cite{li2021uncertainty} (AAAI2022) &ResNet101 & \xmark & 41.5 \\
RIB \cite{lee2021reducing} (NeurIPS2021) & ResNet101 &\xmark& 43.8 \\
\rowcolor{lightgray} GETAM (ours) & ViT-Hybrid &\checkmark& 36.4\\
\bottomrule
\end{tabular}
}
\caption{Comparison with state-of-the-art on MS-COCO \cite{lin2014microsoft}. GETAM is the first end-to-end method evaluated on MS-COCO that performs on par with existing methods.
}
\label{table coco}
\end{table}

\subsection{Comparison to the State-of-the-art Methods}
\subsubsection{PASCAL VOC}
In Table \ref{table sota}, we give a detailed comparison of our proposed approach with other WSSS methods.
%
In the single-stage section, 
we achieve significantly improved performance over all existing end-to-end methods. Comparing to previous state-of-the-art method AALR \cite{zhang2021adaptive}, we have an impressive 7.8\% mIoU increase.
Compared to multi-stage methods, 
our result
based on ViT\_Hybrid (71.7 mIoU on \textit{val} and 72.3 mIoU on \textit{test}) achieves new state-of-the-art on PASCAL VOC.
And our single-stage performances with other transformers including ViT, Deit and Deit\_Distilled are also significantly ahead of existing single-stage approaches, and competitive with multi-stage ones, showing the robustness of GETAM to transformer backbones.
Notably, our approach is the first single-stage method that outperforms the multi-stage methods, which employ sophisticated pipelines and train multiple networks. 
For example, the previous state-of-the-art URN \cite{li2021uncertainty} requires three stages with many inter-dependencies, our single-stage method still outperforms it by 0.5 mIoU on \textit{val} and 0.8 mIoU on \textit{test}.

\subsubsection{MS-COCO}
To further demonstrate the proposed method's effectiveness across different datasets, we evaluate it on the challenging MS-COCO dataset \cite{lin2014microsoft}.
As shown in Table \ref{table coco}, our method achieves 36.4\% mIoU. 
PMM \cite{Li_2021_ICCV}, URN \cite{li2021uncertainty} and RIB \cite{lee2021reducing} are the only methods that outperform ours. They are all multi-stage methods, requiring training at least two networks.
RIB \cite{lee2021reducing} has four training steps to obtain final segmentation results.
Our method still outperforms them on PASCAL VOC (Table \ref{table sota}), which demonstrates promising ability when trained over different datasets.
Notably, our method is the first end-to-end WSSS method that has reported a result on MS-COCO which saves on training complexity.

\section{Conclusion}
In this paper, we propose a vision transformer based WSSS framework by exploring the activation mechanism for transformers. 
Specifically, we propose a new transformer-based activation visualisation approach,  GETAM. It generates  class-wise attention maps that can better capture the object shape compared to previous methods. 
Based on GETAM, we introduce an activation aware label completion module to generate high-quality pseudo labels, it
adopts saliency information 
to refine foreground object masks 
without suppressing background objects.
Finally, we present a novel double-backward propagation scheme to integrate the proposed modules into an end-to-end training framework.
We validate GETAM on weakly the supervised semantic segmentation task,
extensive experimental results on both multi-stage and single-stage training show the effectiveness of our method with different transformer backbones. 
The proposed method offers a new perspective for WSSS using vision transformers,
and we believe that it can further facilitate related research areas.

{\small
\bibliographystyle{ieee_fullname}
\bibliography{egbib}

\begin{thebibliography}{10}\itemsep=-1pt

\bibitem{ahn2019weakly}
Jiwoon Ahn, Sunghyun Cho, and Suha Kwak.
\newblock Weakly supervised learning of instance segmentation with inter-pixel
  relations.
\newblock In {\em IEEE Conference on Computer Vision and Pattern Recognition
  (CVPR)}, pages 2209--2218, Long Beach, CA, USA, 2019. Computer Vision
  Foundation / {IEEE}.

\bibitem{ahn2018learning}
Jiwoon Ahn and Suha Kwak.
\newblock Learning pixel-level semantic affinity with image-level supervision
  for weakly supervised semantic segmentation.
\newblock In {\em IEEE Conference on Computer Vision and Pattern Recognition
  (CVPR)}, pages 4981--4990. Computer Vision Foundation / {IEEE}, 2018.

\bibitem{Araslanov_2020_CVPR}
Nikita Araslanov and Stefan Roth.
\newblock Single-stage semantic segmentation from image labels.
\newblock In {\em IEEE Conference on Computer Vision and Pattern Recognition
  (CVPR)}, pages 4252--4261. Computer Vision Foundation / {IEEE}, June 2020.

\bibitem{bach2015pixel}
Sebastian Bach, Alexander Binder, Gr{\'e}goire Montavon, Frederick Klauschen,
  Klaus-Robert M{\"u}ller, and Wojciech Samek.
\newblock On pixel-wise explanations for non-linear classifier decisions by
  layer-wise relevance propagation.
\newblock {\em PloS one}, 10(7):e0130140, 2015.

\bibitem{bao2021beit}
Hangbo Bao, Li Dong, and Furu Wei.
\newblock Beit: Bert pre-training of image transformers, 2021.

\bibitem{bearman2016s}
Amy Bearman, Olga Russakovsky, Vittorio Ferrari, and Li Fei-Fei.
\newblock What’s the point: Semantic segmentation with point supervision.
\newblock In {\em European Conference on Computer Vision (ECCV)}, pages
  549--565, 2016.

\bibitem{chang2020mixup}
Yu{-}Ting Chang, Qiaosong Wang, Wei{-}Chih Hung, Robinson Piramuthu, Yi{-}Hsuan
  Tsai, and Ming{-}Hsuan Yang.
\newblock Mixup-cam: Weakly-supervised semantic segmentation via uncertainty
  regularization.
\newblock In {\em British Machine Vision Conference (BMVC)}, Virtual Event, UK,
  2020. {BMVA} Press.

\bibitem{chang2020weakly}
Yu-Ting Chang, Qiaosong Wang, Wei-Chih Hung, Robinson Piramuthu, Yi-Hsuan Tsai,
  and Ming-Hsuan Yang.
\newblock Weakly-supervised semantic segmentation via sub-category exploration.
\newblock In {\em IEEE Conference on Computer Vision and Pattern Recognition
  (CVPR)}, pages 8991--9000. Computer Vision Foundation / {IEEE}, 2020.

\bibitem{chattopadhay2018grad}
Aditya Chattopadhay, Anirban Sarkar, Prantik Howlader, and Vineeth~N
  Balasubramanian.
\newblock Grad-cam++: Generalized gradient-based visual explanations for deep
  convolutional networks.
\newblock In {\em 2018 IEEE winter conference on applications of computer
  vision (WACV)}, pages 839--847. IEEE, 2018.

\bibitem{chefer2021generic}
Hila Chefer, Shir Gur, and Lior Wolf.
\newblock Generic attention-model explainability for interpreting bi-modal and
  encoder-decoder transformers, 2021.

\bibitem{chefer2021transformer}
Hila Chefer, Shir Gur, and Lior Wolf.
\newblock Transformer interpretability beyond attention visualization, 2021.

\bibitem{chen2014semantic}
Liang{-}Chieh Chen, George Papandreou, Iasonas Kokkinos, Kevin Murphy, and
  Alan~L. Yuille.
\newblock Semantic image segmentation with deep convolutional nets and fully
  connected crfs.
\newblock In Yoshua Bengio and Yann LeCun, editors, {\em International
  Conference on Learning Representations (ICLR)}, 2015.

\bibitem{chen2017deeplab}
Liang-Chieh Chen, George Papandreou, Iasonas Kokkinos, Kevin Murphy, and Alan~L
  Yuille.
\newblock Deeplab: Semantic image segmentation with deep convolutional nets,
  atrous convolution, and fully connected crfs.
\newblock {\em IEEE Transactions on Pattern Analysis and Machine Intelligence
  (TPAMI)}, 40(4):834--848, 2017.

\bibitem{chen2017rethinking}
Liang-Chieh Chen, George Papandreou, Florian Schroff, and Hartwig Adam.
\newblock Rethinking atrous convolution for semantic image segmentation.
\newblock {\em ArXiv e-prints}, 2017.

\bibitem{dai2015boxsup}
Jifeng Dai, Kaiming He, and Jian Sun.
\newblock Boxsup: Exploiting bounding boxes to supervise convolutional networks
  for semantic segmentation.
\newblock In {\em IEEE International Conference on Computer Vision (ICCV)},
  pages 1635--1643, 2015.

\bibitem{dai2021coatnet}
Zihang Dai, Hanxiao Liu, Quoc Le, and Mingxing Tan.
\newblock Coatnet: Marrying convolution and attention for all data sizes.
\newblock {\em Advances in Neural Information Processing Systems}, 34, 2021.

\bibitem{dosovitskiy2020image}
Alexey Dosovitskiy, Lucas Beyer, Alexander Kolesnikov, Dirk Weissenborn,
  Xiaohua Zhai, Thomas Unterthiner, Mostafa Dehghani, Matthias Minderer, Georg
  Heigold, Sylvain Gelly, et~al.
\newblock An image is worth 16x16 words: Transformers for image recognition at
  scale.
\newblock {\em arXiv preprint arXiv:2010.11929}, 2020.

\bibitem{everingham2010pascal}
Mark Everingham, Luc Van~Gool, Christopher~KI Williams, John Winn, and Andrew
  Zisserman.
\newblock The pascal visual object classes (voc) challenge.
\newblock {\em International Journal of Computer Vision (IJCV)},
  88(2):303--338, 2010.

\bibitem{fan2020learning}
Junsong Fan, Zhaoxiang Zhang, Chunfeng Song, and Tieniu Tan.
\newblock Learning integral objects with intra-class discriminator for
  weakly-supervised semantic segmentation.
\newblock In {\em IEEE Conference on Computer Vision and Pattern Recognition
  (CVPR)}, pages 4283--4292. Computer Vision Foundation / {IEEE}, 2020.

\bibitem{fan2020employing}
Junsong Fan, Zhaoxiang Zhang, and Tieniu Tan.
\newblock Employing multi-estimations for weakly-supervised semantic
  segmentation.
\newblock In {\em European Conference on Computer Vision (ECCV)}, pages
  332--348, 2020.

\bibitem{fan2020cian}
Junsong Fan, Zhaoxiang Zhang, Tieniu Tan, Chunfeng Song, and Jun Xiao.
\newblock Cian: Cross-image afﬁnity net for weakly supervised semantic
  segmentation.
\newblock In {\em AAAI Conference on Artificial Intelligence (AAAI)}, 2020.

\bibitem{gao2021tscam}
Wei Gao, Fang Wan, Xingjia Pan, Zhiliang Peng, Qi Tian, Zhenjun Han, Bolei
  Zhou, and Qixiang Ye.
\newblock Ts-cam: Token semantic coupled attention map for weakly supervised
  object localization, 2021.

\bibitem{guo2019mixup}
Hongyu Guo, Yongyi Mao, and Richong Zhang.
\newblock Mixup as locally linear out-of-manifold regularization.
\newblock In {\em AAAI Conference on Artificial Intelligence (AAAI)},
  volume~33, pages 3714--3722, 2019.

\bibitem{hariharan2011semantic}
Bharath Hariharan, Pablo Arbel{\'a}ez, Lubomir Bourdev, Subhransu Maji, and
  Jitendra Malik.
\newblock Semantic contours from inverse detectors.
\newblock In {\em IEEE International Conference on Computer Vision (ICCV)},
  pages 991--998. IEEE, 2011.

\bibitem{hong2016learning}
Seunghoon Hong, Junhyuk Oh, Honglak Lee, and Bohyung Han.
\newblock Learning transferrable knowledge for semantic segmentation with deep
  convolutional neural network.
\newblock In {\em IEEE Conference on Computer Vision and Pattern Recognition
  (CVPR)}, pages 3204--3212. Computer Vision Foundation / {IEEE}, 2016.

\bibitem{hou2018self}
Qibin Hou, PengTao Jiang, Yunchao Wei, and Ming-Ming Cheng.
\newblock Self-erasing network for integral object attention.
\newblock In {\em Conference on Neural Information Processing Systems
  (NeurIPS)}, pages 549--559, 2018.

\bibitem{huang2018weakly}
Zilong Huang, Xinggang Wang, Jiasi Wang, Wenyu Liu, and Jingdong Wang.
\newblock Weakly-supervised semantic segmentation network with deep seeded
  region growing.
\newblock In {\em IEEE Conference on Computer Vision and Pattern Recognition
  (CVPR)}, pages 7014--7023. Computer Vision Foundation / {IEEE}, 2018.

\bibitem{jiang2019integral}
Peng-Tao Jiang, Qibin Hou, Yang Cao, Ming-Ming Cheng, Yunchao Wei, and Hong-Kai
  Xiong.
\newblock Integral object mining via online attention accumulation.
\newblock In {\em IEEE International Conference on Computer Vision (ICCV)},
  pages 2070--2079, 2019.

\bibitem{jiang2021layercam}
Peng-Tao Jiang, Chang-Bin Zhang, Qibin Hou, Ming-Ming Cheng, and Yunchao Wei.
\newblock Layercam: Exploring hierarchical class activation maps for
  localization.
\newblock {\em IEEE Transactions on Image Processing (TIP)}, 30:5875--5888,
  2021.

\bibitem{kim2021discriminative}
Beomyoung Kim, Sangeun Han, and Junmo Kim.
\newblock Discriminative region suppression for weakly-supervised semantic
  segmentation.
\newblock In {\em AAAI Conference on Artificial Intelligence (AAAI)}, pages
  1754--1761, 2021.

\bibitem{kolesnikov2016seed}
Alexander Kolesnikov and Christoph~H Lampert.
\newblock Seed, expand and constrain: Three principles for weakly-supervised
  image segmentation.
\newblock In {\em European Conference on Computer Vision (ECCV)}, pages
  695--711. Springer, 2016.

\bibitem{Kweon_2021_ICCV}
Hyeokjun Kweon, Sung-Hoon Yoon, Hyeonseong Kim, Daehee Park, and Kuk-Jin Yoon.
\newblock Unlocking the potential of ordinary classifier: Class-specific
  adversarial erasing framework for weakly supervised semantic segmentation.
\newblock In {\em Proceedings of the IEEE/CVF International Conference on
  Computer Vision (ICCV)}, pages 6994--7003, October 2021.

\bibitem{lee2021reducing}
Jungbeom Lee, Jooyoung Choi, Jisoo Mok, and Sungroh Yoon.
\newblock Reducing information bottleneck for weakly supervised semantic
  segmentation.
\newblock {\em Advances in Neural Information Processing Systems}, 34, 2021.

\bibitem{lee2019ficklenet}
Jungbeom Lee, Eunji Kim, Sungmin Lee, Jangho Lee, and Sungroh Yoon.
\newblock Ficklenet: Weakly and semi-supervised semantic image segmentation
  using stochastic inference.
\newblock In {\em IEEE Conference on Computer Vision and Pattern Recognition
  (CVPR)}, pages 5267--5276. Computer Vision Foundation / {IEEE}, 2019.

\bibitem{lee2021anti}
Jungbeom Lee, Eunji Kim, and Sungroh Yoon.
\newblock Anti-adversarially manipulated attributions for weakly and
  semi-supervised semantic segmentation.
\newblock In {\em IEEE Conference on Computer Vision and Pattern Recognition
  (CVPR)}, pages 4071--4080. Computer Vision Foundation / {IEEE}, 2021.

\bibitem{lee2021bbam}
Jungbeom Lee, Jihun Yi, Chaehun Shin, and Sungroh Yoon.
\newblock Bbam: Bounding box attribution map for weakly supervised semantic and
  instance segmentation, 2021.

\bibitem{lee2021railroad}
Seungho Lee, Minhyun Lee, Jongwuk Lee, and Hyunjung Shim.
\newblock Railroad is not a train: Saliency as pseudo-pixel supervision for
  weakly supervised semantic segmentation.
\newblock In {\em IEEE Conference on Computer Vision and Pattern Recognition
  (CVPR)}, pages 5495--5505. Computer Vision Foundation / {IEEE}, 2021.

\bibitem{li2021uncertainty}
Yi Li, Yiqun Duan, Zhanghui Kuang, Yimin Chen, Wayne Zhang, and Xiaomeng Li.
\newblock Uncertainty estimation via response scaling for pseudo-mask noise
  mitigation in weakly-supervised semantic segmentation.
\newblock {\em arXiv preprint arXiv:2112.07431}, 2021.

\bibitem{Li_2021_ICCV}
Yi Li, Zhanghui Kuang, Liyang Liu, Yimin Chen, and Wayne Zhang.
\newblock Pseudo-mask matters in weakly-supervised semantic segmentation.
\newblock In {\em Proceedings of the IEEE/CVF International Conference on
  Computer Vision (ICCV)}, pages 6964--6973, October 2021.

\bibitem{lin2016scribblesup}
Di Lin, Jifeng Dai, Jiaya Jia, Kaiming He, and Jian Sun.
\newblock Scribblesup: Scribble-supervised convolutional networks for semantic
  segmentation.
\newblock In {\em IEEE Conference on Computer Vision and Pattern Recognition
  (CVPR)}, pages 3159--3167. Computer Vision Foundation / {IEEE}, 2016.

\bibitem{lin2014microsoft}
Tsung-Yi Lin, Michael Maire, Serge Belongie, James Hays, Pietro Perona, Deva
  Ramanan, Piotr Doll{\'a}r, and C~Lawrence Zitnick.
\newblock Microsoft coco: Common objects in context.
\newblock In {\em European Conference on Computer Vision (ECCV)}, pages
  740--755, 2014.

\bibitem{liu2021swin}
Ze Liu, Yutong Lin, Yue Cao, Han Hu, Yixuan Wei, Zheng Zhang, Stephen Lin, and
  Baining Guo.
\newblock Swin transformer: Hierarchical vision transformer using shifted
  windows.
\newblock In {\em Proceedings of the IEEE/CVF International Conference on
  Computer Vision}, pages 10012--10022, 2021.

\bibitem{long2015fully}
Jonathan Long, Evan Shelhamer, and Trevor Darrell.
\newblock Fully convolutional networks for semantic segmentation.
\newblock In {\em IEEE Conference on Computer Vision and Pattern Recognition
  (CVPR)}, pages 3431--3440. Computer Vision Foundation / {IEEE}, 2015.

\bibitem{luo2020learning}
Wenfeng Luo and Meng Yang.
\newblock Learning saliency-free model with generic features for
  weakly-supervised semantic segmentation.
\newblock In {\em AAAI Conference on Artificial Intelligence (AAAI)}, pages
  11717--11724, 2020.

\bibitem{mao2021transformer}
Yuxin Mao, Jing Zhang, Zhexiong Wan, Yuchao Dai, Aixuan Li, Yunqiu Lv, Xinyu
  Tian, Deng-Ping Fan, and Nick Barnes.
\newblock Transformer transforms salient object detection and camouflaged
  object detection, 2021.

\bibitem{oh2021background}
Youngmin Oh, Beomjun Kim, and Bumsub Ham.
\newblock Background-aware pooling and noise-aware loss for weakly-supervised
  semantic segmentation.
\newblock In {\em IEEE Conference on Computer Vision and Pattern Recognition
  (CVPR)}, pages 6913--6922. Computer Vision Foundation / {IEEE}, 2021.

\bibitem{papandreou2015weakly}
George Papandreou, Liang-Chieh Chen, Kevin~P Murphy, and Alan~L Yuille.
\newblock Weakly-and semi-supervised learning of a deep convolutional network
  for semantic image segmentation.
\newblock In {\em IEEE International Conference on Computer Vision (ICCV)},
  pages 1742--1750, 2015.

\bibitem{pinheiro2015image}
Pedro~O Pinheiro and Ronan Collobert.
\newblock From image-level to pixel-level labeling with convolutional networks.
\newblock In {\em IEEE Conference on Computer Vision and Pattern Recognition
  (CVPR)}, pages 1713--1721. Computer Vision Foundation / {IEEE}, 2015.

\bibitem{ranftl2021vision}
Ren{\'e} Ranftl, Alexey Bochkovskiy, and Vladlen Koltun.
\newblock Vision transformers for dense prediction.
\newblock In {\em Proceedings of the IEEE/CVF International Conference on
  Computer Vision}, pages 12179--12188, 2021.

\bibitem{roy2017combining}
Anirban Roy and Sinisa Todorovic.
\newblock Combining bottom-up, top-down, and smoothness cues for weakly
  supervised image segmentation.
\newblock In {\em IEEE Conference on Computer Vision and Pattern Recognition
  (CVPR)}, pages 3529--3538. Computer Vision Foundation / {IEEE}, 2017.

\bibitem{ru2022weakly}
Lixiang Ru, Bo Du, Yibing Zhan, and Chen Wu.
\newblock Weakly-supervised semantic segmentation with visual words learning
  and hybrid pooling.
\newblock {\em International Journal of Computer Vision}, pages 1--18, 2022.

\bibitem{selvaraju2017grad}
Ramprasaath~R Selvaraju, Michael Cogswell, Abhishek Das, Ramakrishna Vedantam,
  Devi Parikh, and Dhruv Batra.
\newblock Grad-cam: Visual explanations from deep networks via gradient-based
  localization.
\newblock In {\em Proceedings of the IEEE international conference on computer
  vision}, pages 618--626, 2017.

\bibitem{selvaraju2016grad}
Ramprasaath~R Selvaraju, Abhishek Das, Ramakrishna Vedantam, Michael Cogswell,
  Devi Parikh, and Dhruv Batra.
\newblock Grad-cam: Why did you say that?
\newblock {\em arXiv preprint arXiv:1611.07450}, 2016.

\bibitem{strudel2021segmenter}
Robin Strudel, Ricardo Garcia, Ivan Laptev, and Cordelia Schmid.
\newblock Segmenter: Transformer for semantic segmentation.
\newblock In {\em Proceedings of the IEEE/CVF International Conference on
  Computer Vision}, pages 7262--7272, 2021.

\bibitem{sun2020mining}
Guolei Sun, Wenguan Wang, Jifeng Dai, and Luc Van~Gool.
\newblock Mining cross-image semantics for weakly supervised semantic
  segmentation.
\newblock {\em arXiv preprint arXiv:2007.01947}, 2020.

\bibitem{Sun_2021_ICCV}
Kunyang Sun, Haoqing Shi, Zhengming Zhang, and Yongming Huang.
\newblock Ecs-net: Improving weakly supervised semantic segmentation by using
  connections between class activation maps.
\newblock In {\em Proceedings of the IEEE/CVF International Conference on
  Computer Vision (ICCV)}, pages 7283--7292, October 2021.

\bibitem{sun20203d}
Weixuan Sun, Jing Zhang, and Nick Barnes.
\newblock 3d guided weakly supervised semantic segmentation.
\newblock In {\em Proceedings of the Asian Conference on Computer Vision},
  2020.

\bibitem{sun2022inferring}
Weixuan Sun, Jing Zhang, and Nick Barnes.
\newblock Inferring the class conditional response map for weakly supervised
  semantic segmentation.
\newblock In {\em Proceedings of the IEEE/CVF Winter Conference on Applications
  of Computer Vision}, pages 2878--2887, 2022.

\bibitem{tang2018regularized}
Meng Tang, Federico Perazzi, Abdelaziz Djelouah, Ismail Ben~Ayed, Christopher
  Schroers, and Yuri Boykov.
\newblock On regularized losses for weakly-supervised cnn segmentation.
\newblock In {\em European Conference on Computer Vision (ECCV)}, pages
  507--522, 2018.

\bibitem{touvron2021training}
Hugo Touvron, Matthieu Cord, Matthijs Douze, Francisco Massa, Alexandre
  Sablayrolles, and Herv{\'e} J{\'e}gou.
\newblock Training data-efficient image transformers \& distillation through
  attention.
\newblock In {\em International Conference on Machine Learning (ICML)}, pages
  10347--10357. PMLR, 2021.

\bibitem{vaswani2017attention}
Ashish Vaswani, Noam Shazeer, Niki Parmar, Jakob Uszkoreit, Llion Jones,
  Aidan~N Gomez, {\L}ukasz Kaiser, and Illia Polosukhin.
\newblock Attention is all you need.
\newblock {\em Advances in neural information processing systems}, 30, 2017.

\bibitem{vernaza2017learning}
Paul Vernaza and Manmohan Chandraker.
\newblock Learning random-walk label propagation for weakly-supervised semantic
  segmentation.
\newblock In {\em IEEE Conference on Computer Vision and Pattern Recognition
  (CVPR)}, pages 7158--7166. Computer Vision Foundation / {IEEE}, 2017.

\bibitem{wang2021pvtv2}
Wenhai Wang, Enze Xie, Xiang Li, Deng-Ping Fan, Kaitao Song, Ding Liang, Tong
  Lu, Ping Luo, and Ling Shao.
\newblock Pvtv2: Improved baselines with pyramid vision transformer, 2021.

\bibitem{wang2021weakly}
Xinggang Wang, Jiapei Feng, Bin Hu, Qi Ding, Longjin Ran, Xiaoxin Chen, and
  Wenyu Liu.
\newblock Weakly-supervised instance segmentation via class-agnostic learning
  with salient images.
\newblock In {\em Proceedings of the IEEE/CVF Conference on Computer Vision and
  Pattern Recognition}, pages 10225--10235, 2021.

\bibitem{wang2018weakly}
Xiang Wang, Shaodi You, Xi Li, and Huimin Ma.
\newblock Weakly-supervised semantic segmentation by iteratively mining common
  object features.
\newblock In {\em IEEE Conference on Computer Vision and Pattern Recognition
  (CVPR)}, pages 1354--1362. Computer Vision Foundation / {IEEE}, 2018.

\bibitem{wang2020self}
Yude Wang, Jie Zhang, Meina Kan, Shiguang Shan, and Xilin Chen.
\newblock Self-supervised equivariant attention mechanism for weakly supervised
  semantic segmentation.
\newblock In {\em IEEE Conference on Computer Vision and Pattern Recognition
  (CVPR)}, pages 12275--12284. Computer Vision Foundation / {IEEE}, 2020.

\bibitem{wei2021shallow}
Jun Wei, Qin Wang, Zhen Li, Sheng Wang, S~Kevin Zhou, and Shuguang Cui.
\newblock Shallow feature matters for weakly supervised object localization.
\newblock In {\em Proceedings of the IEEE/CVF Conference on Computer Vision and
  Pattern Recognition}, pages 5993--6001, 2021.

\bibitem{wu2021cvt}
Haiping Wu, Bin Xiao, Noel Codella, Mengchen Liu, Xiyang Dai, Lu Yuan, and Lei
  Zhang.
\newblock Cvt: Introducing convolutions to vision transformers.
\newblock In {\em Proceedings of the IEEE/CVF International Conference on
  Computer Vision}, pages 22--31, 2021.

\bibitem{wu2021embedded}
Tong Wu, Junshi Huang, Guangyu Gao, Xiaoming Wei, Xiaolin Wei, Xuan Luo, and
  Chi~Harold Liu.
\newblock Embedded discriminative attention mechanism for weakly supervised
  semantic segmentation.
\newblock In {\em IEEE Conference on Computer Vision and Pattern Recognition
  (CVPR)}, pages 16765--16774. Computer Vision Foundation / {IEEE}, 2021.

\bibitem{Xu_2021_ICCV}
Lian Xu, Wanli Ouyang, Mohammed Bennamoun, Farid Boussaid, Ferdous Sohel, and
  Dan Xu.
\newblock Leveraging auxiliary tasks with affinity learning for weakly
  supervised semantic segmentation.
\newblock In {\em Proceedings of the IEEE/CVF International Conference on
  Computer Vision (ICCV)}, pages 6984--6993, October 2021.

\bibitem{yao2021non}
Yazhou Yao, Tao Chen, Guo-Sen Xie, Chuanyi Zhang, Fumin Shen, Qi Wu, Zhenmin
  Tang, and Jian Zhang.
\newblock Non-salient region object mining for weakly supervised semantic
  segmentation.
\newblock In {\em IEEE Conference on Computer Vision and Pattern Recognition
  (CVPR)}, pages 2623--2632. Computer Vision Foundation / {IEEE}, 2021.

\bibitem{yun2019cutmix}
Sangdoo Yun, Dongyoon Han, Seong~Joon Oh, Sanghyuk Chun, Junsuk Choe, and
  Youngjoon Yoo.
\newblock Cutmix: Regularization strategy to train strong classifiers with
  localizable features.
\newblock In {\em IEEE International Conference on Computer Vision (ICCV)},
  pages 6023--6032, 2019.

\bibitem{zeng2019joint}
Yu Zeng, Yunzhi Zhuge, Huchuan Lu, and Lihe Zhang.
\newblock Joint learning of saliency detection and weakly supervised semantic
  segmentation.
\newblock In {\em IEEE International Conference on Computer Vision (ICCV)},
  pages 7223--7233, 2019.

\bibitem{zhang2020reliability}
Bingfeng Zhang, Jimin Xiao, Yunchao Wei, Mingjie Sun, and Kaizhu Huang.
\newblock Reliability does matter: An end-to-end weakly supervised semantic
  segmentation approach.
\newblock In {\em AAAI Conference on Artificial Intelligence (AAAI)}, pages
  12765--12772, 2020.

\bibitem{zhang2020causal}
Dong Zhang, Hanwang Zhang, Jinhui Tang, Xiansheng Hua, and Qianru Sun.
\newblock Causal intervention for weakly-supervised semantic segmentation.
\newblock {\em arXiv preprint arXiv:2009.12547}, 2020.

\bibitem{Zhang_2021_ICCV}
Fei Zhang, Chaochen Gu, Chenyue Zhang, and Yuchao Dai.
\newblock Complementary patch for weakly supervised semantic segmentation.
\newblock In {\em Proceedings of the IEEE/CVF International Conference on
  Computer Vision (ICCV)}, pages 7242--7251, October 2021.

\bibitem{zhang2020splitting}
Tianyi Zhang, Guosheng Lin, Weide Liu, Jianfei Cai, and Alex Kot.
\newblock Splitting vs. merging: Mining object regions with discrepancy and
  intersection loss for weakly supervised semantic segmentation.
\newblock In {\em European Conference on Computer Vision (ECCV)}, 2020.

\bibitem{zhang2021adaptive}
Xiangrong Zhang, Zelin Peng, Peng Zhu, Tianyang Zhang, Chen Li, Huiyu Zhou, and
  Licheng Jiao.
\newblock Adaptive affinity loss and erroneous pseudo-label refinement for
  weakly supervised semantic segmentation.
\newblock In {\em Proceedings of the 29th ACM International Conference on
  Multimedia}, pages 5463--5472, 2021.

\bibitem{zhang2021aggregating}
Zizhao Zhang, Han Zhang, Long Zhao, Ting Chen, and Tomas Pfister.
\newblock Aggregating nested transformers.
\newblock In {\em arXiv preprint arXiv:2105.12723}, 2021.

\bibitem{zhao2017pyramid}
Hengshuang Zhao, Jianping Shi, Xiaojuan Qi, Xiaogang Wang, and Jiaya Jia.
\newblock Pyramid scene parsing network.
\newblock In {\em IEEE Conference on Computer Vision and Pattern Recognition
  (CVPR)}, pages 2881--2890. Computer Vision Foundation / {IEEE}, 2017.

\bibitem{zhou2016learning}
Bolei Zhou, Aditya Khosla, Agata Lapedriza, Aude Oliva, and Antonio Torralba.
\newblock Learning deep features for discriminative localization.
\newblock In {\em IEEE Conference on Computer Vision and Pattern Recognition
  (CVPR)}, pages 2921--2929. Computer Vision Foundation / {IEEE}, 2016.

\end{thebibliography}
}


\end{document}